\documentclass[10pt,twocolumn,letterpaper]{article}

\usepackage{3dv}
\usepackage{times}
\usepackage{epsfig}
\usepackage{graphicx}
\usepackage{amsmath}
\usepackage{amssymb}
\usepackage{ tabularx, wrapfig}
\usepackage{amsthm, textcomp, mathtools}
\usepackage{enumitem}
\usepackage{subcaption}
\usepackage{booktabs}
\usepackage{array}
\usepackage{multirow}
\usepackage{enumitem}
\usepackage{wasysym}
\usepackage{color}
\usepackage[pagebackref=true,breaklinks=true,letterpaper=true,colorlinks,bookmarks=false]{hyperref}

\newcolumntype{L}[1]{>{\raggedright\let\newline\\\arraybackslash\hspace{0pt}}p{#1}}
\newcolumntype{C}[1]{>{\centering\let\newline\\\arraybackslash\hspace{0pt}}b{#1}}
\newcolumntype{R}[1]{>{\raggedleft\let\newline\\\arraybackslash\hspace{0pt}}b{#1}}
\definecolor{darkblue}{rgb}{0.2, 0.3, 0.53}
\definecolor{wine}{rgb}{0.5, 0.04, 0.1}

\hypersetup{
     colorlinks   = true,
     citecolor    = darkblue,
     linkcolor    = wine
}

\threedvfinalcopy 

\def\threedvPaperID{84} 

\ifthreedvfinal\fi
\begin{document}

\title{Learning a Hierarchical Latent-Variable Model of 3D Shapes}
\author{Shikun Liu\\
Imperial College London\\
{\tt\small shikun.liu17@imperial.ac.uk}
\and
C. Lee Giles \\
Pennsylvania State University\\
{\tt\small giles@ist.psu.edu}
\and
Alexander G. Ororbia II\\
Pennsylvania State University\\
{\tt\small ago109@psu.edu }
}

\twocolumn[{%
\renewcommand\twocolumn[1][]{#1}%
\maketitle
\vspace{-1.5em}
\begin{center}
  \begin{tabular}{c}
      \includegraphics[width=0.1\linewidth]{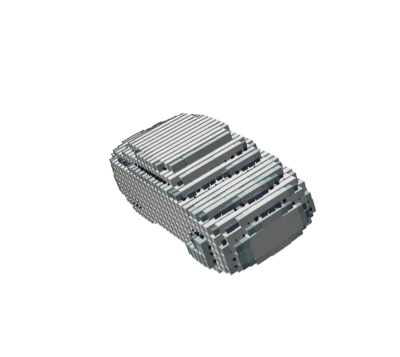}
      \includegraphics[width=0.1\linewidth]{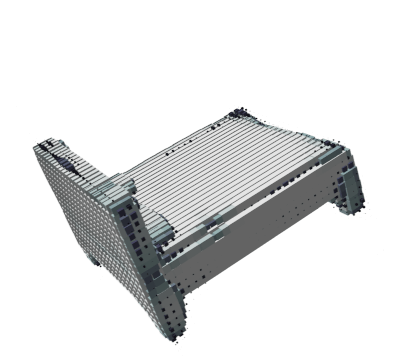}
      \includegraphics[width=0.1\linewidth]{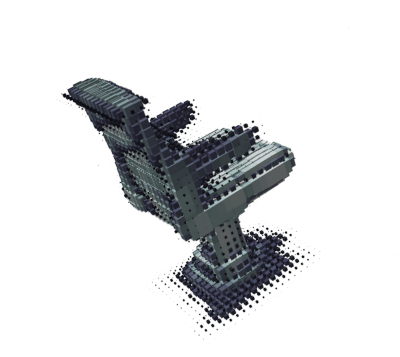}
      \includegraphics[width=0.1\linewidth]{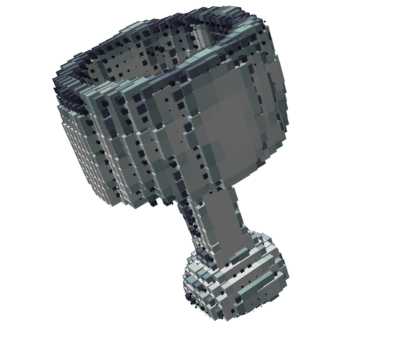}
      \includegraphics[width=0.1\linewidth]{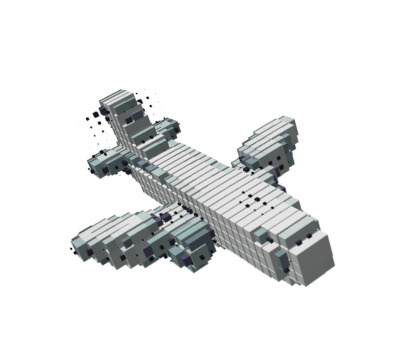}
      \includegraphics[width=0.1\linewidth]{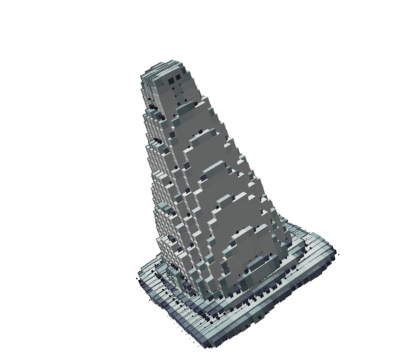}
      \includegraphics[width=0.1\linewidth]{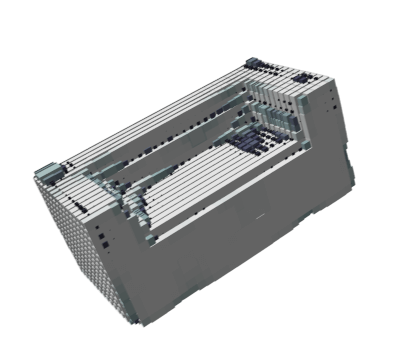}
      \includegraphics[width=0.1\linewidth]{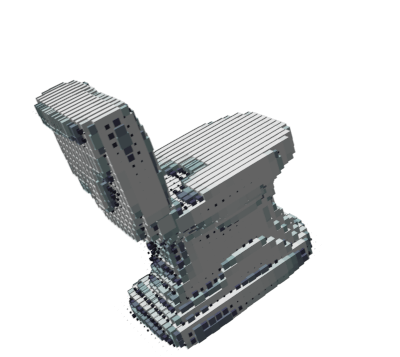}
      \includegraphics[width=0.1\linewidth]{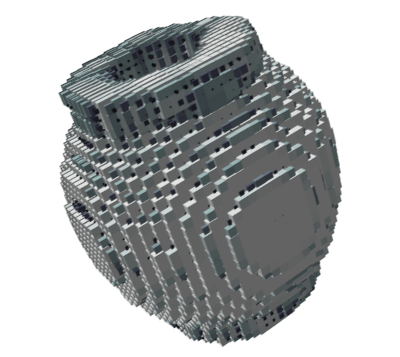}
    \end{tabular}
    \captionof{figure}{Random 3D shapes generated by sampling the learned latent space of the proposed Variational Shape Learner trained from ModelNet40 dataset \cite{wu20153d}. See more results in Section \ref{subsec:shapelearn}.}
    \label{fig:randomgen}
    \vspace{1em}
\end{center}%
}]

\begin{abstract}
  We propose the Variational Shape Learner (VSL), a generative model that learns the underlying structure of voxelized 3D shapes in an unsupervised fashion. Through the use of skip-connections, our model can successfully learn and infer a latent, hierarchical representation of objects. Furthermore, realistic 3D objects can be easily generated by sampling the VSL's latent probabilistic manifold. We show that our generative model can be trained end-to-end from 2D images to perform single image 3D model retrieval. Experiments show, both quantitatively and qualitatively, the improved generalization of our proposed model over a range of tasks, performing better or comparable to various state-of-the-art alternatives.
\end{abstract}

\section{Introduction}
\label{sec:intro}
Over the past several years, impressive strides have been made in the generative modelling of 3D objects. Much of this progress can be attributed to recent advances in artificial neural network research.  Instead of the usual approach to representing 3D shapes with voxel occupancy vectors, promising recent work has taken to learning simple latent representations of such objects. Neural architectures that have been developed with this goal in mind include those based on deep belief networks \cite{wu20153d}, deep autoencoders \cite{zhu2016deep,girdhar2016learning,rezende2016unsupervised}, and 3D convolutional networks \cite{maturana2015voxnet,yan2016perspective,sedaghat2016orientation,choy20163d,hane2017hierarchical}. The positive progress made so far with neural networks has also led to the creation of several large-scale 3D CAD model benchmarks, notably ModelNet \cite{wu20153d} and ShapeNet \cite{chang2015shapenet}.

However, despite the progress made so far, one key weakness shared among all previous state-of-the-art approaches is that all of them have focused on learning a single (``flat'') vector representation of 3D shapes. These include recent and powerful models such as the autoencoder-like T-L Network  \cite{girdhar2016learning} and the probabilistic 3D Generative Adversarial Network (3D-GAN) \cite{wu2016learning}, which shared its latent vector representation over multiple tasks. Other models, such as those of \cite{kar2015category,kar2017learning}, further required additional supervision using information about camera viewpoints, shape keypoints, and segmentations.

To describe the input with only a single layer of latent variables might be too restrictive an assumption, hindering the expressiveness of the underlying generative model learned. Having a multilevel latent structure, on the other hand, would allow for lower-level latent variables to focus on modelling features such as edges and the upper levels to learn to command those lower-level variables as to where to place those edges in order to form curves and shapes. This composition of latent (local) sub-structures would allow us to exploit the fact that most 3D shapes usually have similar structure. Note that this course-to-fine feature extraction process is the very essence of abstraction, yielding representations that can be easily constructed in terms of less abstract ones \cite{bengio2013deep}. Higher-level variables, or disentangled features, would be modelling complex interactions of low-level patterns. Thus, to encourage the learning of hierarchical features, we explicitly incorporate this as a prior in our model through explicit architectural constraints.

In this paper, motivated by the argument for a hierarchical representation developed above and the promise shown in work such as that of \cite{edwards2016towards}, we show how to encourage a latent-variable generative model to learn a hierarchy of latent variables through the use of synaptic skip-connections. These skip-connections encourage each layer of latent variables to model exactly one level of abstraction of the data. To efficiently learn such a latent structure, we further exploit recent advances in approximate inference \cite{kingma2013auto} to develop a variational learning procedure. Empirically, we show that the learned generative model, which we call the Variational Shape Learner, acquires rich representations of 3D shapes that yield significantly improved performance across a multitude of 3D shape tasks.

The main contributions of this paper are as follows:
\begin{itemize}[leftmargin=*]
  \item We propose a novel latent-variable model, which we call the Variational Shape Learner, which is capable of learning expressive feature representations of 3D shapes. We observe impressive performance in shape generation and shape arithmetic in a large dataset.
  \item For both general 3D model building and single image reconstruction, we show that our model is fully unsupervised, requiring no extra human-generated information about segmentation, keypoints, or pose information.
  \item We show that our model outperforms current state-of-the-art approaches in unsupervised (object) model classification while requiring significantly fewer learned feature extractors (a vector with less than 100 dimensions compared to the 3D-GAN's 2.5 million dimensional vector).
  \item In real-world image reconstruction, our extensive set of experiments show that the proposed Variational Shape Learner surpasses state-of-the-art in 8 of 10 classes. Half of these the VSL surpasses by a large margin.
\end{itemize}

\section{Related Work} 
\label{sec:rw}
3D object recognition is a well-studied problem in the computer vision literature. Early efforts \cite{patterson2008object,knopp2010hough,rusu2009fast} often combined simple image classification methods with hand-crafted shape descriptors, requiring intensive effort on the side of the human data annotator. However, ever since the ImageNet contest of 2012 \cite{krizhevsky2012imagenet}, deep convolutional networks (ConvNets) \cite{fukushima1988neocognitron,lecun1989backpropagation} have swept the vision industry, becoming nearly ubiquitous in countless applications.

Research in learning probabilistic generative models has also benefited from the advances made by artificial neural networks.  Generative Adversarial Networks (GANs), proposed in \cite{goodfellow2014generative} and Variational auto-encoders (VAEs), proposed in \cite{kingma2013auto,rezende2014stochastic}, are some of the most popular and important frameworks that have emerged from improvements in generative modelling. Successful adaptation of these frameworks range from a focus in natural language and speech processing \cite{chung2015recurrent,serban2017piecewise} to realistic image synthesis \cite{gregor2015draw,radford2015unsupervised,pu2016variational}, yielding promising, positive results. Nevertheless, very little work, outside of \cite{wu2016learning,girdhar2016learning,rezende2016unsupervised}, has focused on modeling 3D objects, where generative architectures could be used to learn probabilistic embeddings. The model proposed in this paper will offer another step towards constructing powerful generative models of 3D structures.

One study, amidst the rise of neural network-based approaches to 3D object recognition, that is most relevant to this paper is that of \cite{wu20153d}, which presented promising results and a benchmark for 3D model recognition: ModelNet. Following this key study, researchers have tried applying 3D ConvNets \cite{maturana2015voxnet,choy20163d,su2015multi,yan2016perspective}, autoencoders \cite{xie2015deepshape,zhu2016deep,girdhar2016learning,rezende2016unsupervised}, and a variety of probabilistic neural generative models \cite{wu2016learning,rezende2016unsupervised} to the problem of 3D model recognition, with each study progressively advancing state-of-the-art.

With respect to 3D object generation from 2D images, commonly used methods can be roughly grouped into two categories: 3D voxel prediction \cite{wu20153d,wu2016learning,girdhar2016learning,rezende2016unsupervised,choy20163d,hane2017hierarchical} and mesh-based methods \cite{gargallo2007minimizing,delaunoy2008minimizing}. The 3D-R2N2 model \cite{choy20163d} represents a more recent approach to the task, which involves training a recurrent neural network to predict 3D voxels from one or more 2D images. \cite{rezende2016unsupervised} also takes a recurrent network-based approach, but receives a depth image as input rather than normal 2D images. The learnable stereo system \cite{kar2017learning} processes one or more camera views and camera pose information to produce compelling 3D object samples.

Many of the above methods require multiple images and/or additional human-provided information. Some approaches have attempted to minimize human involvement by developing weakly-supervised schemes, making use of image silhouettes to conduct 3D object reconstruction \cite{yan2016perspective,Wiles2017SilNetS}. Of the few unsupervised neural-based approaches that exist, the T-L network \cite{girdhar2016learning} is one of the most important, combining a convolutional autoencoder with an image regressor to encode a unified vector representation of a given 2D image. However, one fundamental issue with the T-L Network is its three-phase training procedure, since jointly training the system components proves to be too difficult. The 3D-GAN \cite{wu2016learning} offers a way to train 3D object models in an adversarial learning scheme. However, GANs are notoriously difficult to train \cite{arjovsky2017towards}, often due to ill-designed loss functions and the higher chance of zero gradients.

In contrast to prior work, our approach, which is derived from a variational Bayesian perspective view of learning, naturally allows for joint training of all model parameters. Furthermore, our approach makes use of a well-formulated loss function that circumvents the instability involved with adversarial learning while still being able to produce higher-quality samples.

\section{The Variational Shape Learner}
\label{sec:arch}
\begin{figure*}[ht!]
  \centering
  \includegraphics[width=0.95\linewidth]{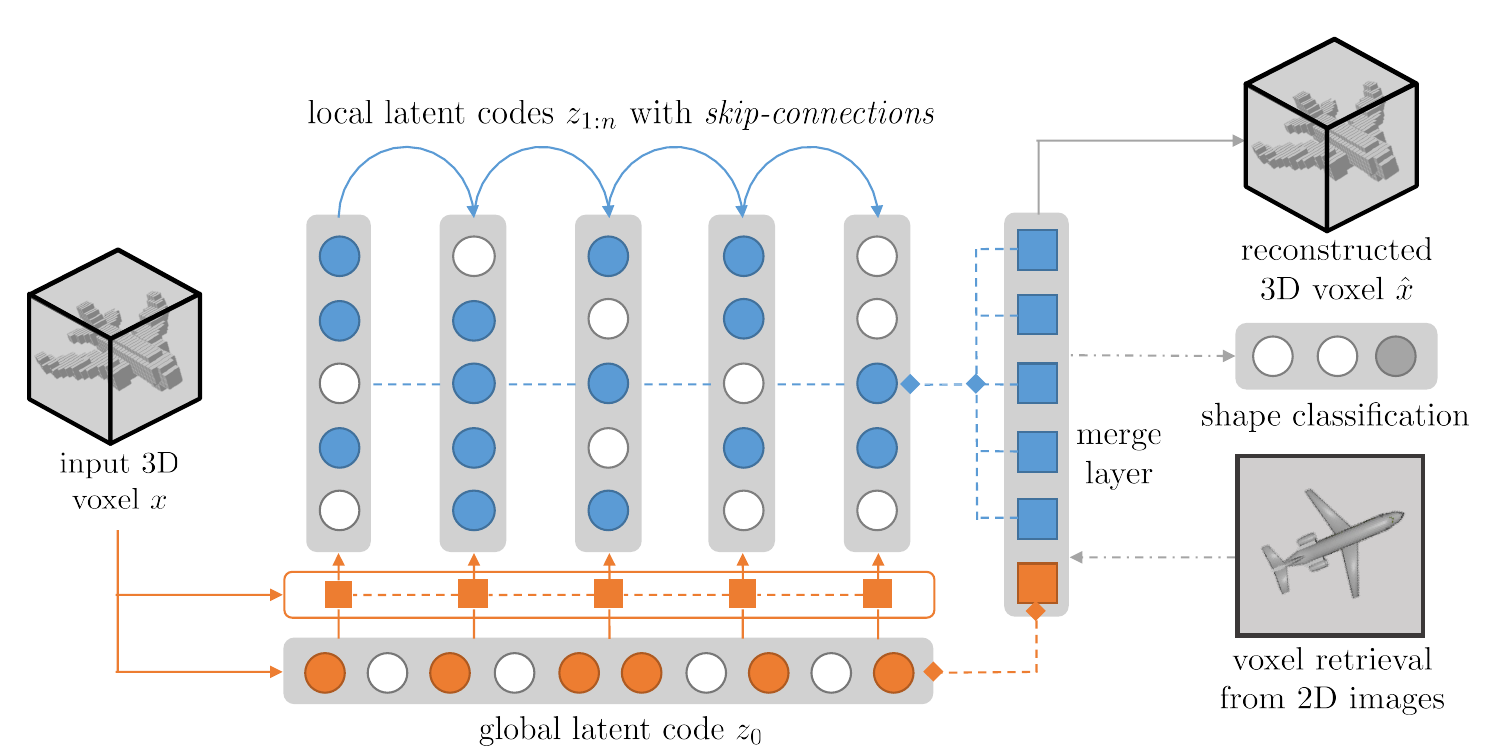}
  \caption{The network structure of the Variational Shape Learner. Solid lines represent synaptic connections for either fully-connected or convolutional layers while dashed lines represent concatenation. Dotted-dashed lines represent possible applications. $\circ$ means latent features, $\Box$ means concatenated features, and $\diamond$ means equivalence relation.}
  \label{fig:vsl}
\end{figure*}

In this section, we introduce our proposed model, the Variational Shape Learner (VSL), which builds on the ideas of the Neural Statistician  \cite{edwards2016towards} and the volumetric convolutional network \cite{maturana2015voxnet}, the parameters of which the VSL learns under a variational inference scheme \cite{kingma2013auto}.

\subsection{The Design Philosophy}
\label{subsec: design}
It is well known that generative models, learned through variational inference, are excellent at reconstructing complex data but tend to produce blurry samples. This happens because there is uncertainty in the model's predictions when we reconstruct the data from a latent space. As described above, previous approaches to 3D object modelling have focused on learning a single latent representation of the data. However, this simple latent structure might be hindering the model's ability to extract richer structure from the input distribution and thus lead to blurrier reconstructions.

To improve the quality of the samples of generated objects, we introduce a more complex internal variable structure, with the specific goal of encouraging the learning of a hierarchical arrangement of latent feature detectors. The motivation for a latent hierarchy comes from the observation that objects under the same category usually have similar geometric structure. As can be seen in Figure \ref{fig:vsl}, we start from a global latent variable layer (horizontally depicted) that is hardwired to a set of local latent variables layers (vertically depicted), each tasked with representing one level of feature abstraction. The skip-connections tie together the latent codes, and in a top-down directed fashion, local codes closer to the input will tend to represent lower-level features while local codes farther away from the input will tend towards representing higher-level features.

The global latent vector can be thought of as a large pool of command units that ensures that each local code extracts information relative to its position in the hierarchy, forming an overall coherent structure. This explicit global-local form, and the way it constrains how information flows across it, lends itself to a straightforward parametrization of the generative model and furthermore ensures robustness, dramatically cutting down on over-fitting.
To make things easier for training via stochastic back-propagation, the local codes will be concatenated to a flattened structure when fed into the task-specific models, e.g., a shape classifier or a voxel reconstruction module.
Ultimately, more realistic samples should be generated by an architecture supporting this kind of latent-variable design, since the local variable layers will robustly encode hierarchical semantic cues in an unsupervised fashion.

\subsection{Model Objective: Variational + Latent Loss}
The variational auto-encoder (VAE) \cite{kingma2013auto} has recently been introduced as a powerful generative model for unsupervised learning. The  generative model $p_\theta(x|z)$ for a single data point $x$ with a latent variable $z$ can be parameterized by a neural network with parameters $\theta$. The parameters are inferred by maximizing the variational lower bound,
\begin{equation}
\label{eq: vae}
    \log p(x) \geq \mathbb{E}_{{q}_\phi(z|x)}\left[\log \frac{p_\theta(x|z)p_\theta(z)}{q_\phi(z|x)} \right]
\end{equation}
The inference model $q_\phi(z|x)$ can also be parameterized by a deep neural network. The inference and generative parameters are then jointly trained by optimizing Equation\ \ref{eq: vae} using back-propagation and stochastic gradient ascent. To deal with the stochasticity of the latent variables, which, in VAE models, are typically assumed to be Gaussian distributed, we use the re-parameterization trick in order to back-propagate through the operation of sampling the Gaussian variables. We refer the reader to \cite{doersch2016tutorial} for a much more detailed explanation.

To learn the parameters of the VSL latent-variable model, we will take a variational inference approach, where the goal is to learn a  generative model $p(x;\theta)$, with generative parameters $\theta$, using a recognition model $q(z_{0:n}|x;\phi)$, with variational parameters $\phi$. The VSL's learning objective contains a standard reconstruction loss term $\mathcal{L}_{rec}$ as well as a regularization penalty $\mathcal{L}_{reg}$ over the latent variables. Furthermore, the loss contains an additional term for the latent variables $\mathcal{L}_{lat}$, which is particularly relevant and useful for the 3D model retrieval task of Section \ref{subsec:imrec}. This extra term is a simple $\mathcal{L}_2$ penalty imposed on the difference between the learned features of the image regressor $z'$ and true latent features $z=[z_{0:n}]$ where $[\cdot]$ denotes concatenation.

We assume a fixed, spherical unit Gaussian prior, $ p(z_0)=\mathcal{N}(0,I)$. The conditional distribution over each local latent code ($z_i,i\geq 2$) is defined as follows:
\begin{equation}
p(z_i|z_{i-1},z_0;\theta)=\mathcal{N}(\mu(z_{i-1},z_0),\sigma^2(z_{i-1},z_0))
\end{equation}
where the first local code $z_1$ is simply:
\begin{equation}
p(z_1|z_0;\theta)=\mathcal{N}(\mu(z_0),\sigma^2(z_0)).
\end{equation}
Note that $p(z_1|z_0;\theta)$ and $p(z_i|z_{i-1},z_0;\theta)$ are also spherical Gaussians and $\theta$ contains the generative parameters.
The (occupancy) probability for one voxel $p(x)$ can then be calculated by,
\begin{equation}
\int p(x|z_{0:n};\theta)p(z_1|z_0;\theta)p(z_0)\prod_{i=2}^n p(z_i|z_{i-1},z_0;\theta)\, dz_{0:n}.
\end{equation}

Let the reconstructed voxel $\hat{x}$ be directly parameterized by occupancy probability. The loss $\mathcal{L}(x)$ for the input voxel $x$ of the VSL is then calculated by the following equation:
\begin{equation}
\label{eq:loss}
\mathcal{L}(x)=\mathcal{L}_{rec}+\delta\mathcal{L}_{reg}+\gamma \mathcal{L}_{lat},
\end{equation}
where each term in the equation above is defined as follows:
\begin{align}
\mathcal{L}_{rec} &= x\log(\hat{x})+(1-x)\log(1-\hat{x}) \\
\begin{split}
\mathcal{L}_{reg} &= \text{{\scshape kl}}(q(z_0|x;\phi)\|p(z_0))\\
&+\text{{\scshape kl}}(q(z_1|z_0,x;\phi)\|p(z_1|z_0;\theta)) \\
&+\sum_{i=2}^n\text{{\scshape kl}}(q(z_i|z_{i-1},z_0,x;\phi)\|p(z_i|z_{i-1},z_0;\theta))
\end{split}
\\
\mathcal{L}_{lat} &= -\|z'-z \|^2_2.
\end{align}
Note that $\delta$ and $\gamma$, which weigh the contributions of the each term towards the overall cost, are tunable hyper-parameters.

\subsection{Encoder: 3D-ConvNet + Skip-Connections}
\label{subsec:encoder}
The global latent code $z_0$ is directly learned from the input voxel through three convolutional layers with kernel sizes $\left\{6,5,4\right\}$, strides $\left\{2,2,1 \right\}$ and channels $\left\{32,64,128\right\}$.

Each local latent code $z_{>1}$ is conditioned on the global latent code, the input voxel $x$, and the previous latent code (except for $z_1$, which does not have a previous latent code) using two fully-connected layers with $100$ neurons each. The skip-connections between local codes help ease the process of learning hierarchical features (i.e., improved gradient transmission) and force each local code to learn one level of abstraction.

The approximate posterior for a single voxel is given by:
\begin{equation}
      q(z_0|x;\phi)q(z_1|z_0,x;\phi) \prod_{i=2}^n q(z_i|z_{i-1},z_0,x;\phi)
\end{equation}
where $\phi$, the variational parameters, is parameterized by neural networks. $n$ represents the number of local latent codes.

\subsection{Decoder: 3D-DeConvNet}
\label{subsec:decoder}
After we learn the global and local latent codes $z_{0:n}$, we then concatenate them into a single vector as shown in Figure \ref{fig:vsl} in blue dashed lines.

A 3D deconvolutional neural network with dimensions symmetrical to the encoder of Section \ref{subsec:encoder} is used to decode the learned latent features into a voxel. An element-wise logistic sigmoid is applied to the output layer in order to convert the learned features to occupancy probabilities for each voxel cell.

\subsection{Image Regressor: 2D-ConvNet}
We use a standard 2D convolutional network to encode input RGB images into a feature space with the same dimension as the concatenation of global and local latent codes $[z_{0:n}]$. The network contains four fully-convolutional layers with kernel sizes $\left\{32, 15, 5, 3 \right\}$, strides $\left\{2,2,2,1 \right\}$, and channels $\left\{16, 32, 64, 128  \right\}$. The last convolutional layer is flattened and fed into two fully-connected layers with 200 and 100 neurons each. Unlike the encoder described in Section \ref{subsec:encoder}, we apply dropout \cite{srivastava2014dropout} before the last fully-connected layer.

\section{Experiments}
\label{sec:3dshape}
To evaluate the quality of our proposed generative model for 3D shapes, we conduct several extensive experiments.

In Section \ref{subsec:shapelearn}, we investigate our model's ability to generalize and synthesize through a shape interpolation experiment and an nearest neighbours analysis of random generated samples from the VSL. Following this, in Section \ref{subsec:shapeclass}, we evaluate our model on the task of unsupervised shape classification by directly using the learned latent features on both the ModelNet10 and ModelNet40 datasets. We compare these results to previous supervised and unsupervised state-of-the-art methods. Next, we test our model's ability to reconstruct real-world image in Section \ref{subsec:imrec}, comparing our results to 3D-R2N2 \cite{choy20163d} and NRSfM \cite{kar2015category}. Finally, we demonstrate the richness of the VSL's learned semantic embeddings through vector arithmetic, using the latent features trained on ModelNet40 for Section \ref{subsec:shapeplus}.

\subsection{Datasets}
\label{subsec:datasets}
{\bf ModelNet} There are two variants of the ModelNet dataset, ModelNet10 and ModelNet 40, introduced in \cite{wu20153d}, with 10 and 40 target classes respectively. ModelNet10 has 3D shapes which are pre-aligned with the same pose across all categories. In contrast, ModelNet40 (which includes the shapes found in ModelNet10) features a variety of poses. We voxelize both ModelNet10 and ModelNet40 with resolution $[30\times 30 \times 30]$. To test our model's ability to handle 3D shapes of great variety and complexity, we use ModelNet40 for most of the experiments, especially for those in Section \ref{subsec:shapelearn} and \ref{subsec:shapeplus}. Both ModelNet10 and ModelNet40 are used to conduct the shape classification experiments.

{\bf PASCAL 3D} The PASCAL 3D dataset is composed of the images from the PASCAL VOC 2012 dataset \cite{everingham2015pascal}, augmented with 3D annotations using PASCAL 3D+ \cite{xiang_wacv14}. We voxelize the 3D CAD models using resolution $[30\times 30 \times 30]$ and use the same training and testing splits of \cite{kar2015category}, which was also used in \cite{choy20163d} to conduct real-world image reconstruction (of which the experiment in Section \ref{subsec:imrec} is based off of). We use the bounding box information as provided in the dataset. Note that the only pre-processing we applied was image cropping and padding with 0-intensity pixels to create final samples of resolution $[100\times 100]$ (which was required for our model).

\subsection{Training Protocol}
\label{subsec:training}
Training was the same across all experiments, with only minor details that were task-dependent.
The architecture of the VSL experimented with in this paper consisted of 5 local latent codes, each made up of 10 variables for ModelNet40 and 5 for ModelNet10. For ModelNet40, the global latent code was set to a dimensionality of 20 variables, while for ModelNet10, it was set to 10 variables.

The hyper-parameter $\delta$ was set to $10^{-3}$ across training on both ModelNet10 and ModelNet40. We optimise parameters by maximizing the loss function defined in Equation \ref{eq:loss} using the Adam adaptive learning rate \cite{kingma2014adam}, with step size set to $5\times 10^{-5}$. For the experiments of Sections \ref{subsec:shapelearn}, \ref{subsec:shapeclass}, and \ref{subsec:shapeplus}, over 2500 epochs, parameter updates were calculated using mini-batches of 200 samples on ModelNet40 and 100 samples on ModelNet10.

For the experiment in Section \ref{subsec:imrec}, we use 5 local latent codes (each with dimensionality of 5) and a global latent code of 20 variables for the jointly trained model. For the separately trained model, we use 3 local latent codes, each with dimensionality of 2, and a global latent code of dimensionality 5. Mini-batches of 40 samples were use to compute gradients for the joint model while 5 samples were used for the separately trained model. For both model variants, dropout \cite{srivastava2014dropout} was to control for over-fitting, with $p_{drop} = 0.2$, and early stopping was employed (resulting in only 150 epochs).

\begin{figure}[ht!]
	\centering
  \begin{subfigure}[b]{0.23\textwidth}
    \centering
    \includegraphics[width=\linewidth]{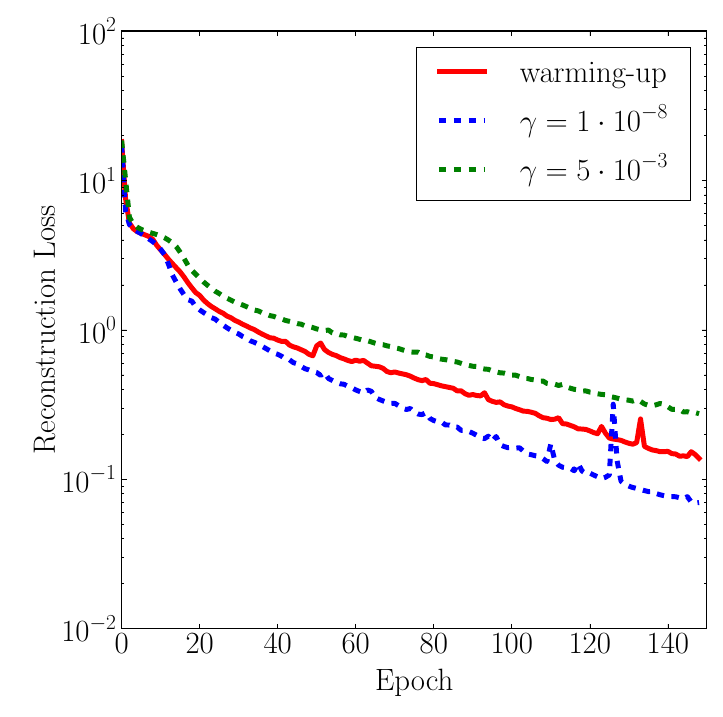}
    \end{subfigure}
  \,
  \begin{subfigure}[b]{0.23\textwidth}
    \centering
    \includegraphics[width=\linewidth]{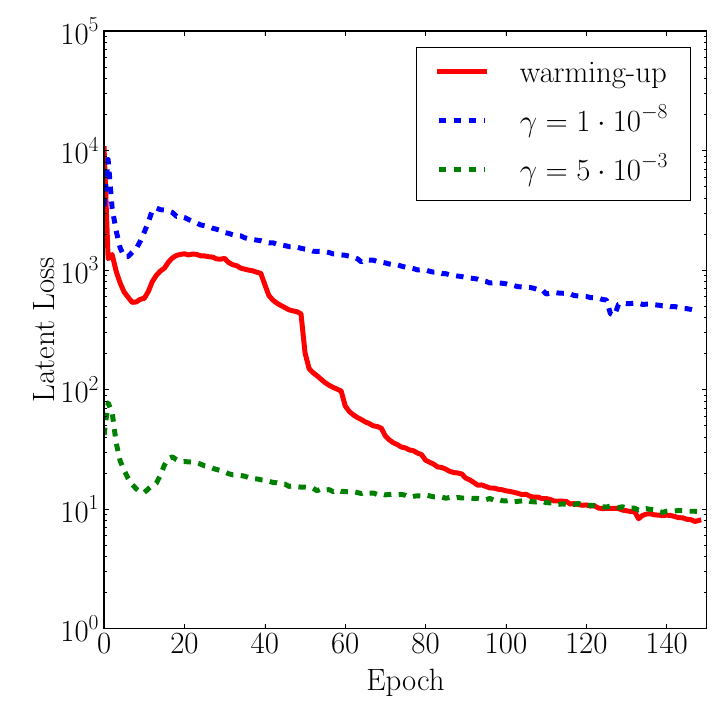}
  \end{subfigure}
	\caption{Training the VSL for image reconstruction using a warming-up schedule compared to using constant weights $\gamma=1\cdot 10^{-8}$ and $\gamma = 5\cdot 10^{-3}$.}
  \label{fig:losscompare}
\end{figure}

For Section \ref{subsec:imrec}, which involved image reconstruction and thus required the loss term $\mathcal{L}_{lat}$, instead of searching for an optimal value of the hyper-parameter $\gamma$ through cross-validation, we employed a ``warming-up'' schedule, similar to that of \cite{sonderby2016ladder}. ``Warming-up'' involves gradually increasing  $\gamma$ (on a log-scale as depicted in Figure \ref{fig:losscompare}), which controls the relative weighting of $\mathcal{L}_{lat}$ in Equation \ref{eq:loss}. The schedule is defined as follows,
\begin{equation}
\gamma =
\begin{dcases}
10^{\lfloor t/10\rfloor -8}\quad  &\phantom{50<{}}t\leq 50\\
\lfloor\frac{t-40}{10}\rfloor\cdot 10^{-3} \quad & 50<t<100\\
5 \cdot 10^{-3}   \quad &\phantom{50<{}} t\geq 100.
\end{dcases}
\end{equation}

Figure \ref{fig:losscompare} depicts, empirically, the benefits of employing a warming-up schedule over using a fixed, externally set coefficient for the $\mathcal{L}_{lat}$ term in our image reconstruction experiment. We remark that using a warming-up schedule plays an essential role in acquiring good performance on the image reconstruction task.

\subsection{Shape Generation and Learning}
\label{subsec:shapelearn}
\begin{figure*}[ht!]
  \centering
  \small
  \setlength\tabcolsep{6pt}
  \begin{tabular}{C{0.08\textwidth}|m{0.1\textwidth}m{0.1\textwidth}m{0.1\textwidth}m{0.1\textwidth}m{0.1\textwidth}m{0.1\textwidth} | m{0.1\textwidth}}
   \multicolumn{1}{C{0.1\textwidth}}{ }  & \multicolumn{6}{c}{Shape Generation}   & \multicolumn{1}{C{0.1\textwidth}}{Nearest Neighbor} \\
 \toprule
  {\it airplane} &
    \includegraphics[width=\linewidth]{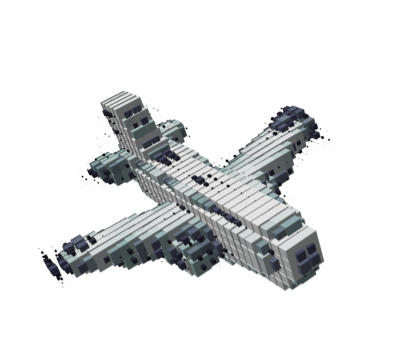}     &
    \includegraphics[width=\linewidth]{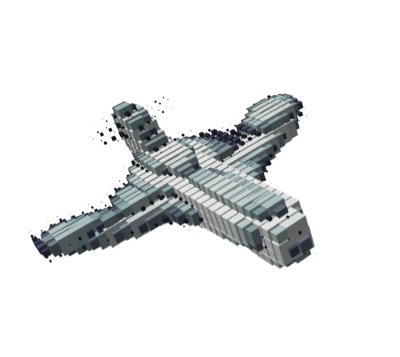}     &
    \includegraphics[width=\linewidth]{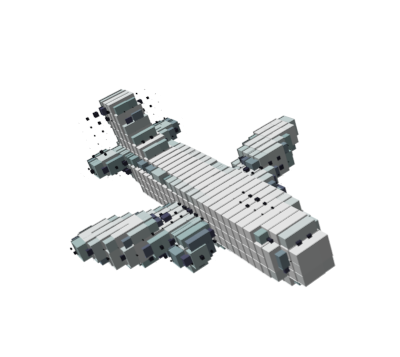}     &
    \includegraphics[width=\linewidth]{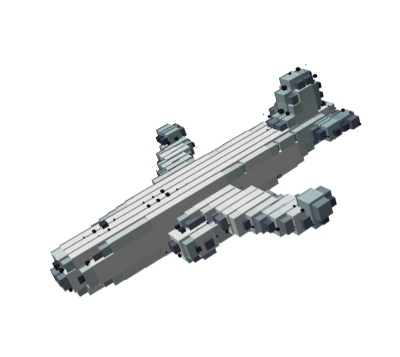}     &
    \includegraphics[width=\linewidth]{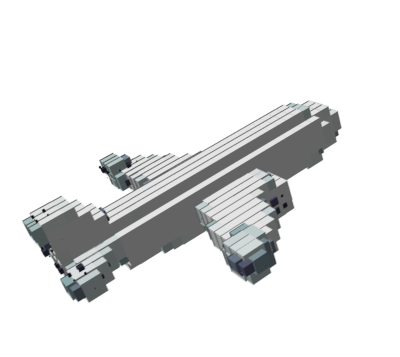}     &
    \includegraphics[width=\linewidth]{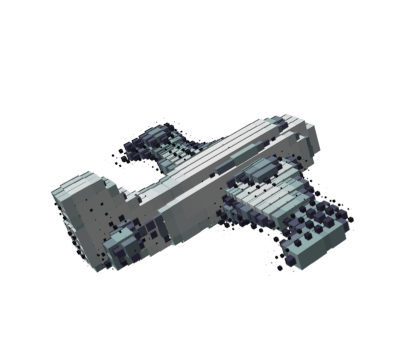}     &
    \includegraphics[width=\linewidth]{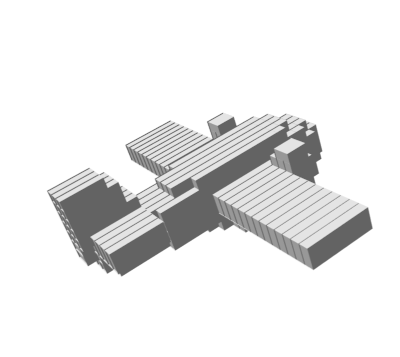}
    \\
    {\it chair} &
     \includegraphics[width=\linewidth]{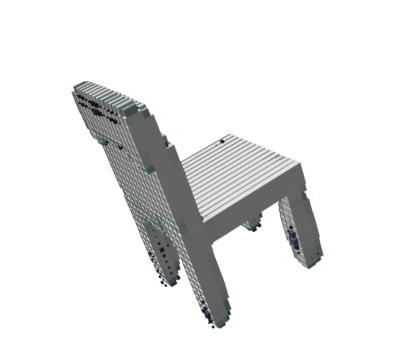}    &
     \includegraphics[width=\linewidth]{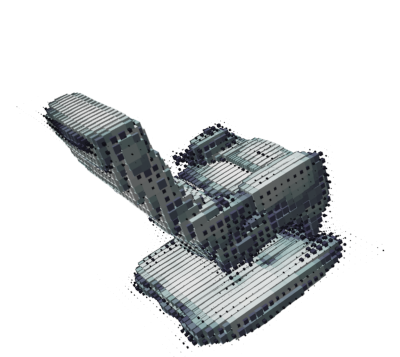}    &
     \includegraphics[width=\linewidth]{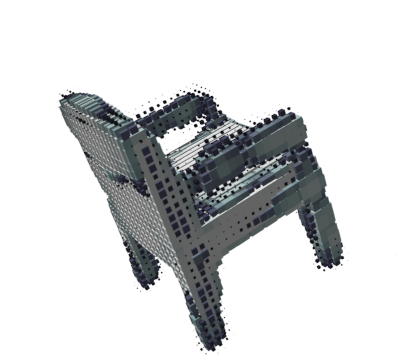}    &
     \includegraphics[width=\linewidth]{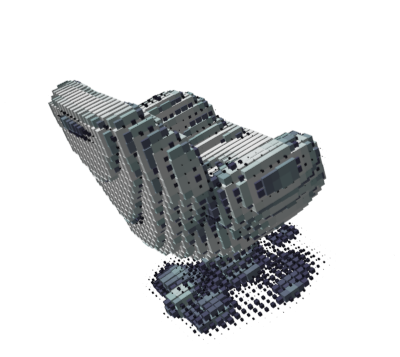}    &
    \includegraphics[width=\linewidth]{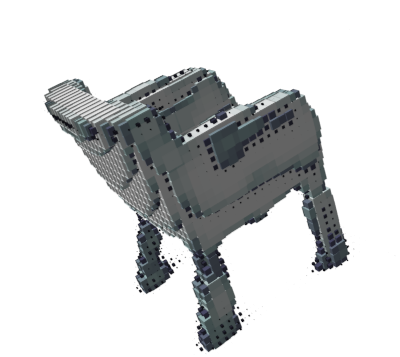}    &
    \includegraphics[width=\linewidth]{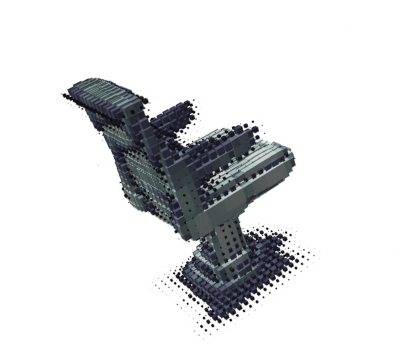}    &
    \includegraphics[width=\linewidth]{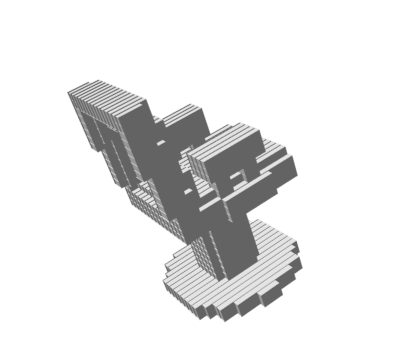}
    \\
    {\it toilet} &
    \includegraphics[width=\linewidth]{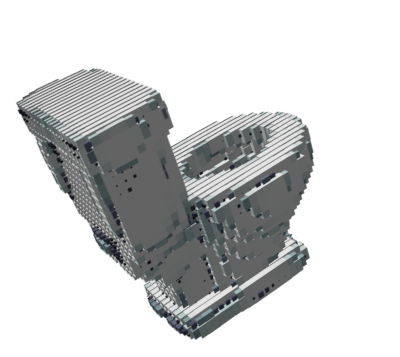}    &
    \includegraphics[width=\linewidth]{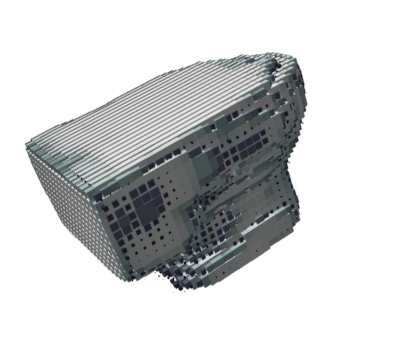}    &
    \includegraphics[width=\linewidth]{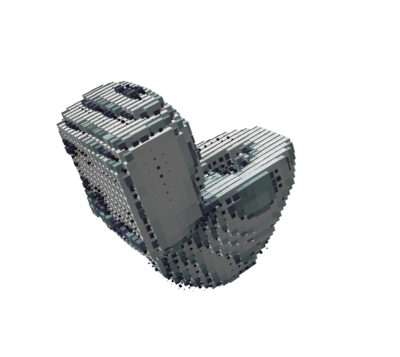}    &
    \includegraphics[width=\linewidth]{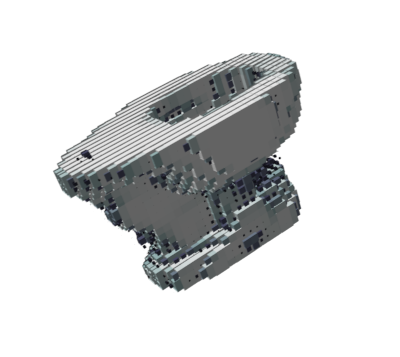}    &
    \includegraphics[width=\linewidth]{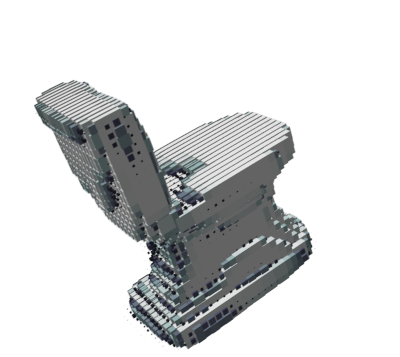}    &
    \includegraphics[width=\linewidth]{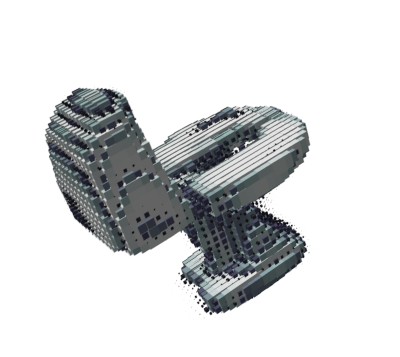}    &
    \includegraphics[width=\linewidth]{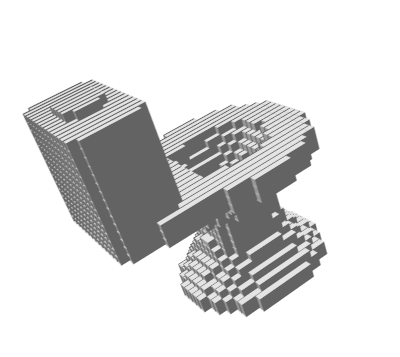}
    \\
    {\it vase}  &
    \includegraphics[width=\linewidth]{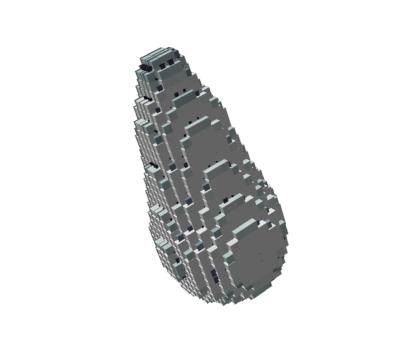}    &
    \includegraphics[width=\linewidth]{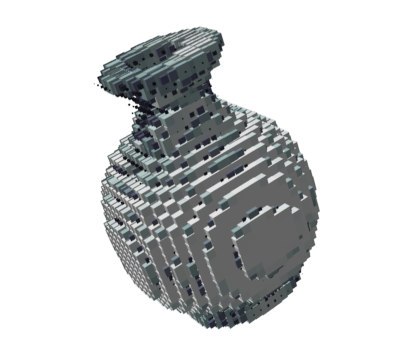}    &
     \includegraphics[width=\linewidth]{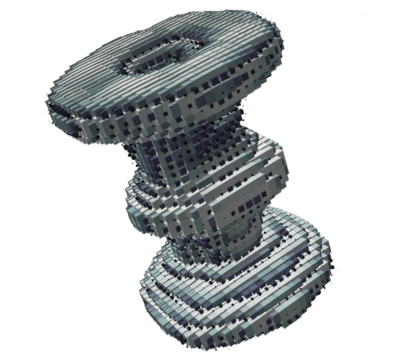}   &
     \includegraphics[width=\linewidth]{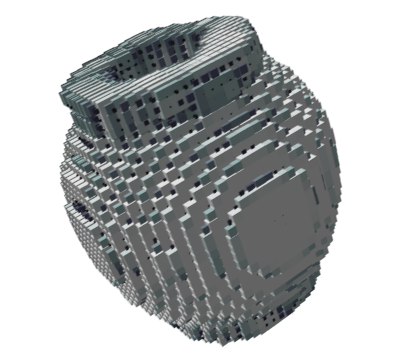}   &
     \includegraphics[width=\linewidth]{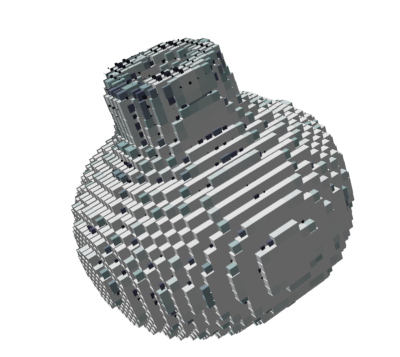}   &
     \includegraphics[width=\linewidth]{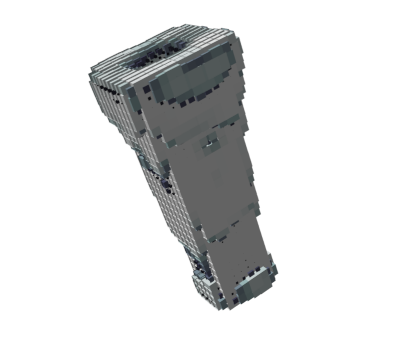}   &
     \includegraphics[width=\linewidth]{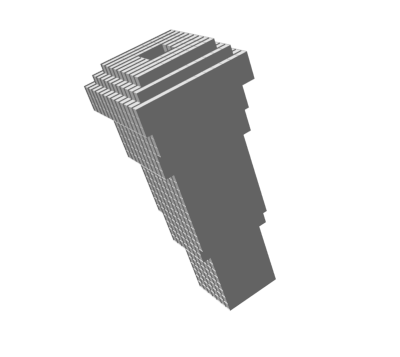}\\
    {\it desk}  &
     \includegraphics[width=\linewidth]{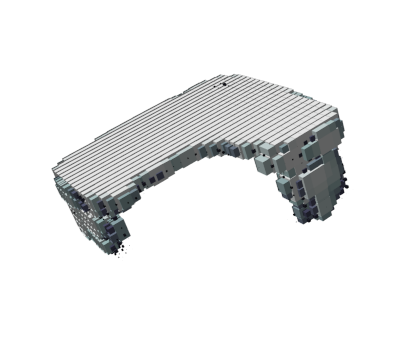}    &
     \includegraphics[width=\linewidth]{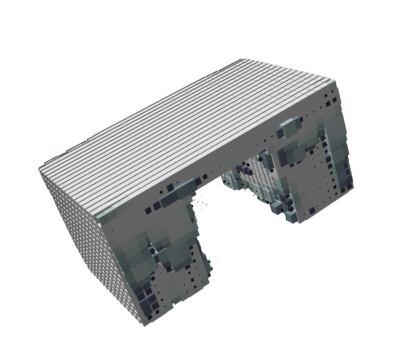}    &
     \includegraphics[width=\linewidth]{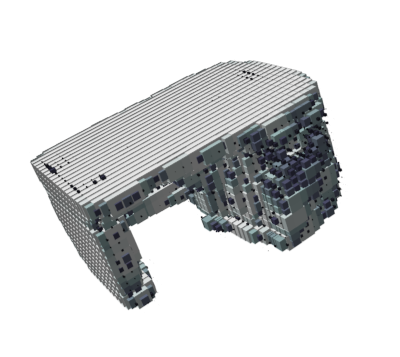}    &
     \includegraphics[width=\linewidth]{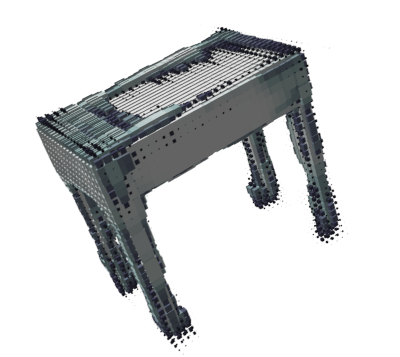}    &
     \includegraphics[width=\linewidth]{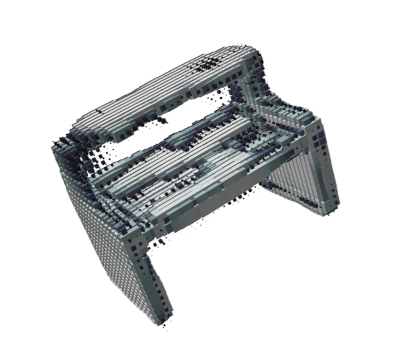}    &
     \includegraphics[width=\linewidth]{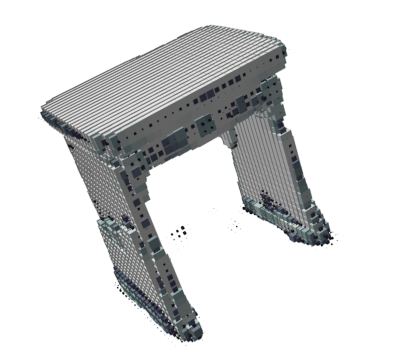}    &
     \includegraphics[width=\linewidth]{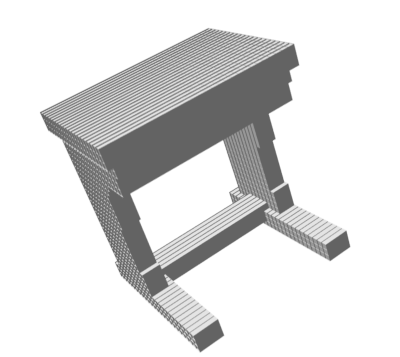}\\
    {\it sofa}  &
    \includegraphics[width=\linewidth]{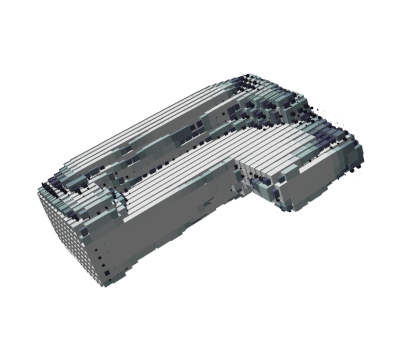}   &
    \includegraphics[width=\linewidth]{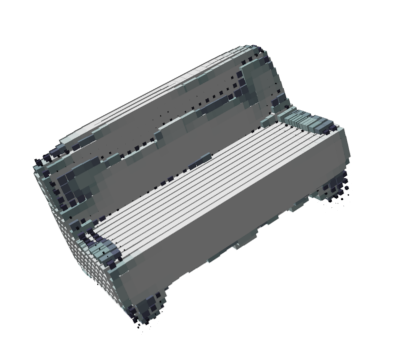}   &
    \includegraphics[width=\linewidth]{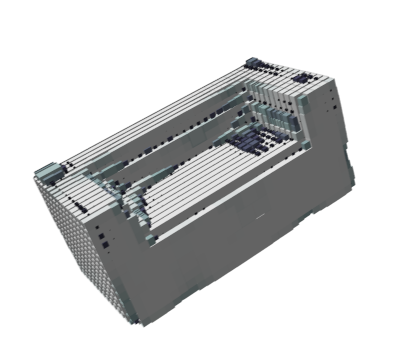}   &
    \includegraphics[width=\linewidth]{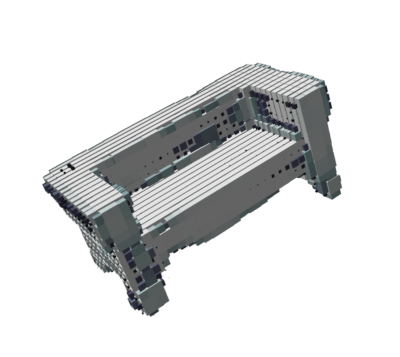}   &
    \includegraphics[width=\linewidth]{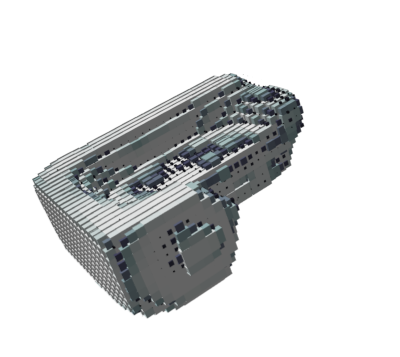}   &
    \includegraphics[width=\linewidth]{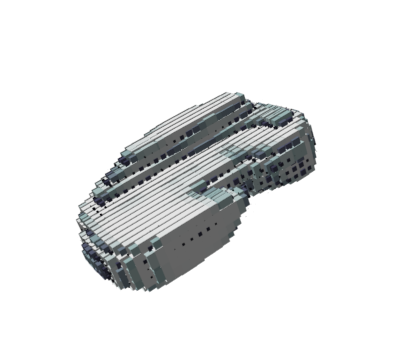}  &
    \includegraphics[width=\linewidth]{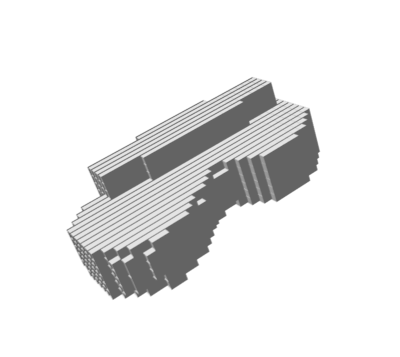}
    \end{tabular}
    \caption{Randomly generated results from the proposed Variational Shape Learner trained on ModelNet40. The nearest neighbors are the ground-truth shapes, fetched from the test data, and placed for reference in the last column of the table. }
    \label{fig:shapegen}
\end{figure*}

\begin{figure*}[ht!]
  \vspace{-0.5em}
  \centering
  \begin{subfigure}[b]{\textwidth}
    \centering
    \includegraphics[width=0.1\linewidth]{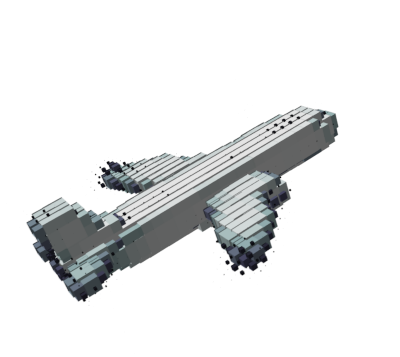}\qquad%
    \includegraphics[width=0.1\linewidth]{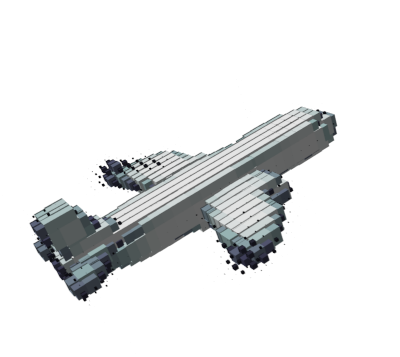}\qquad%
    \includegraphics[width=0.1\linewidth]{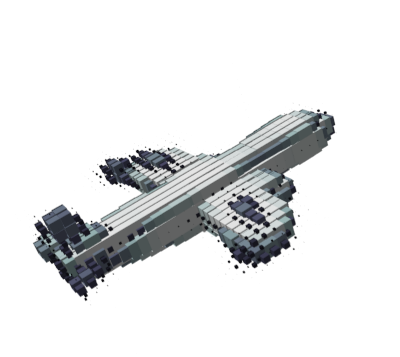}\qquad%
    \includegraphics[width=0.1\linewidth]{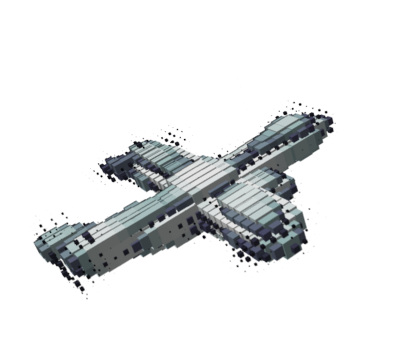}\qquad%
    \includegraphics[width=0.1\linewidth]{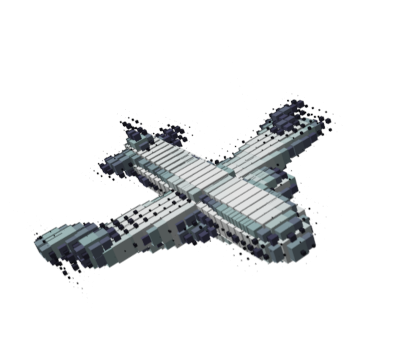}\qquad%
    \includegraphics[width=0.1\linewidth]{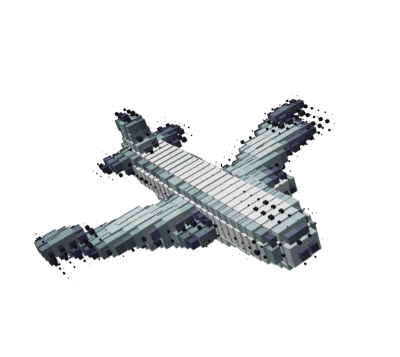}\qquad%
    \includegraphics[width=0.1\linewidth]{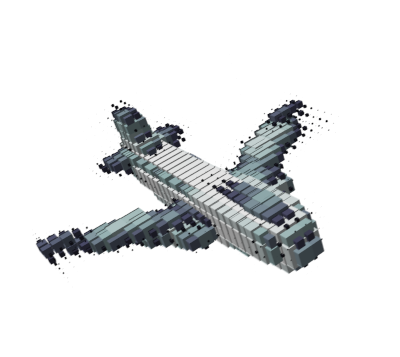}\qquad%
    \caption*{\it Intra-Class Interpolation (airplane)}
  \end{subfigure}
  \quad
  \begin{subfigure}[b]{\textwidth}
    \centering
    \includegraphics[width=0.1\linewidth]{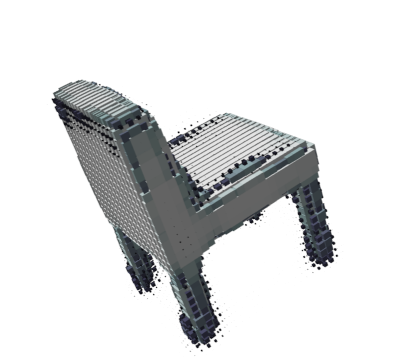}\qquad%
    \includegraphics[width=0.1\linewidth]{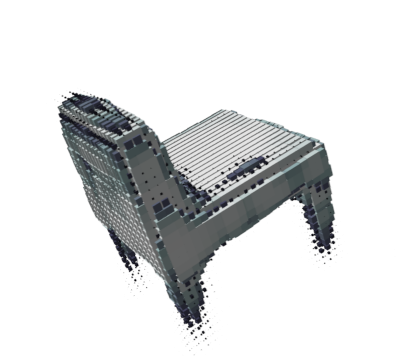}\qquad%
    \includegraphics[width=0.1\linewidth]{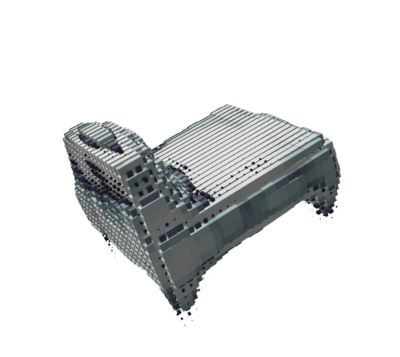}\qquad%
    \includegraphics[width=0.1\linewidth]{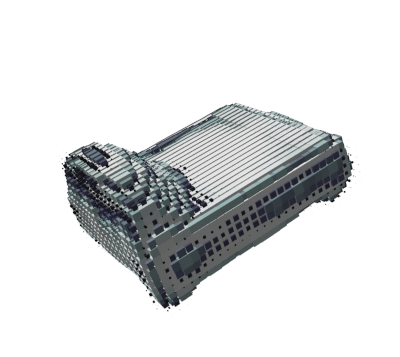}\qquad%
    \includegraphics[width=0.1\linewidth]{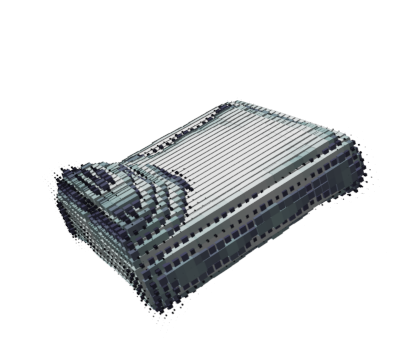}\qquad%
    \includegraphics[width=0.1\linewidth]{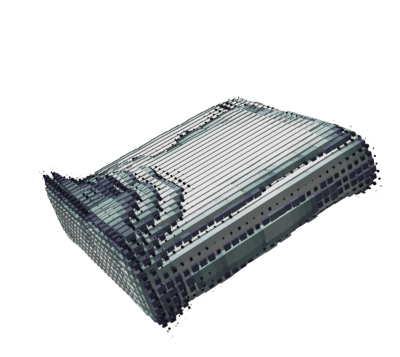}\qquad%
    \includegraphics[width=0.1\linewidth]{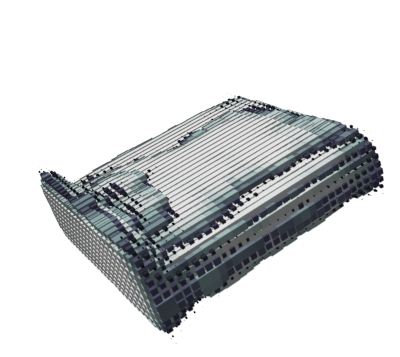}\qquad%
    \caption*{\it Inter-Class Interpolation (chair $\to$ bed)}
  \end{subfigure}
  \caption{Interpolation results of the Variational Shape Learner on ModelNet40.}
  \label{fig:shapeinterp}
\end{figure*}

\begin{figure*}[ht!]
  \centering
  \setlength\tabcolsep{13pt}
    \begin{tabular}{C{0.07\textwidth}m{0.07\textwidth}m{0.07\textwidth}m{0.07\textwidth}C{0.07\textwidth}m{0.07\textwidth}m{0.07\textwidth}m{0.07\textwidth}}
      {\it airplane}    &  \includegraphics[width=\linewidth]{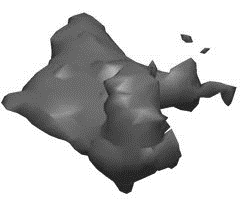}  &  \includegraphics[width=\linewidth]{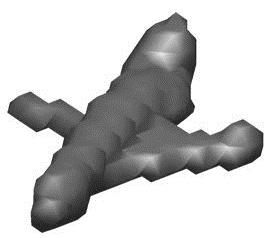}  &  \includegraphics[width=\linewidth]{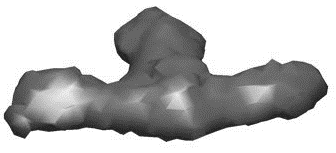}  &
      {\it desk}   &      \includegraphics[width=\linewidth]{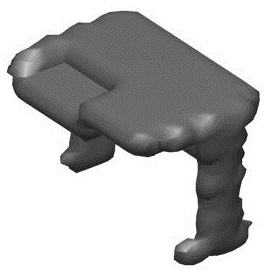}   &  \includegraphics[width=\linewidth]{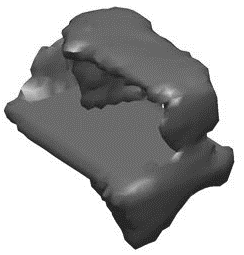}   &  \includegraphics[width=\linewidth]{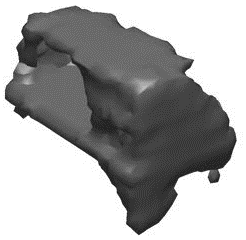}  \\
      \midrule
      {\it sofa}    &  \includegraphics[width=\linewidth]{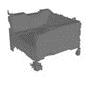}  &  \includegraphics[width=\linewidth]{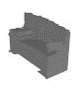}  &  \includegraphics[width=\linewidth]{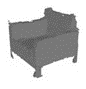}  &
      {\it chair}   &      \includegraphics[width=0.8\linewidth]{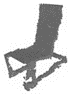}   &  \includegraphics[width=0.7\linewidth]{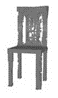}   &  \includegraphics[width=0.8\linewidth]{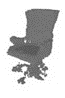}  \\
    \end{tabular}
    \vspace{-0.5em}
    \caption{Shape generation from previous state-of-the-art approaches. Up: generated shapes in resolution $[30\times 30\times 30]$ from \cite{wu20153d}; Down: generated shapes in resolution $[64\times 64\times 64]$ from \cite{wu2016learning}.}
  \label{fig:shapecompare}%
\end{figure*}

To examine our model's ability to generate high-resolution 3D shapes with realistic details, we design a task that involves shape generation and shape interpolation. We add Gaussian noise to the learned latent codes on test data taken from ModelNet40 and then use our model to generate ``unseen'' samples that are similar to the input voxel. In effect, we generate objects from our VSL model directly from vectors, without a reference object/image.

The results of our shape interpolation experiment, from both within-class and across-class perspectives, is presented in Figure \ref{fig:shapeinterp}. It can be observed that the proposed VSL shows the ability to smoothly transition between two objects. Our results on shape generation are shown in Figure \ref{fig:shapegen}. Notably, in our visualizations, darker colours correspond to smaller occupancy probability while lighter corresponds to higher occupancy probability. We further compare to previous state-of-the-art results in shape generation, which are depicted in Figure \ref{fig:shapecompare}.

During training, we observed that our model was robust to different choices of the number and dimensionality of its local/global latent codes. We provide the table below as an ablative analysis showing how test reconstruction error is affected by various settings of the latent variables. From the results, we can observe a clear trend that the network with higher dimensionality and greater number of latent variables tends to generate better results. However, increasing the number of network parameters to attain better accuracy also brings about slower training, an important trade-off that one will need to consider in various application scenarios.

\begin{table}[ht!]
  \centering
  \small
  \setlength\tabcolsep{2pt}
        \def\arraystretch{0.9}
    \begin{tabular}{ccccc}
    \multicolumn{3}{c}{\bf Latent Space Parameters} & \multicolumn{2}{c}{\bf Reconstruction Error} \\
    Global Dim. & Local Dim. & Local Num. & Model-10 & Model-40 \\
    \midrule
    10    & 5     & 3     & 0.0931       & 0.0860 \\
    10    & 5     & 5     & 0.0903      & 0.0831\\
    20    & 10    & 5     & 0.0907      & 0.0798 \\
    50    & 10    & 10    & 0.0910      &  0.0767\\
    100   & 20    & 10    & 0.0902      &  0.0789\\
    \end{tabular}
    \caption{Reconstruction error of ModelNet 10/40 with various choices of network structure. }
\end{table}
\vspace{-0.5em}

\subsection{Shape Classification}
\label{subsec:shapeclass}
One way to test model expressiveness is to conduct shape classification directly using the learned embeddings. We evaluate the features learned on the ModelNet dataset \cite{wu20153d} by concatenating both the global latent variable with the local latent layers, creating a single feature vector $[z_{0:m}]$. We train a Support Vector Machine with an RBF kernel for classification using these ``pre-trained'' embeddings.

\begin{table}[ht!]
 \vspace{-0.5em}
  \centering
  \small
  \setlength\tabcolsep{2pt}
    \begin{tabular}{clcc}
    \multirow{2}[0]{*}{{\bf Supervision}} & \multicolumn{1}{c}{\multirow{2}[0]{*}{{\bf Method}}} & \multicolumn{2}{c}{{\bf Classification Rate}} \\
          &       & ModelNet10 & ModelNet40 \\
    \toprule
    \multirow{6}[0]{*}{Supervised} & 3D ShapeNets \cite{wu20153d}  & 83.5\% & 77.3\% \\
          & DeepPano \cite{shi2015deeppano} & 85.5\% &  77.6\% \\
          & Geometry Image \cite{sinha2016deep} & 88.4\% & 83.9\% \\
          & VoxNet \cite{maturana2015voxnet}& 92.0\%  & 83.0\%   \\
          & PointNet \cite{qi2016pointnet} & - & 89.2\% \\
          & ORION \cite{sedaghat2016orientation} &   93.8\%     &  - \\
    \midrule
    \multirow{6}[0]{*}{Unsupervised} & SPH \cite{kazhdan2003rotation}  &  79.8\%  & 68.2\% \\
          & LFD \cite{chen2003visual}  &   79.9\%    & 75.5\% \\          & T-L Network \cite{girdhar2016learning}&   74.4\%  & - \\
          & VConv-DAE \cite{sharma2016vconv} & 80.5\%  &    75.5\%   \\
          & 3D-GAN \cite{wu2016learning}&  91.0\% & 83.3\% \\
          & VSL (ours) &  {\bf 91.0}\%   & {\bf 84.5}\%\\
    \bottomrule
    \end{tabular}%
 \caption{ModelNet classification results for both unsupervised and supervised methods.}
  \label{tab:shapeclass}%
   \vspace{-0.5em}
\end{table}%

Table \ref{tab:shapeclass} shows the performance of previous state-of-the-art supervised and unsupervised methods in shape classification on both variants of the ModelNet dataset. Notably, the best unsupervised state-of-the-art results reported so far were from the 3D-GAN of \cite{wu2016learning}, which used features from 3 layers of convolutional networks with total dimensions $[62\times 32^3 + 128\times 16^3 + 56\times 8^3]$. This is a far larger feature space than that required by our model, which is simply $[5\times 5 + 10]$ (for 10-way classification) and $[5\times 10 +20]$ (for 40-way classification) and reaches the exact same level of performance. The VSL performs comparably to supervised state-of-the-art, outperforming models such as 3D ShapeNet \cite{wu20153d}, DeepPano \cite{shi2015deeppano}, and the geometry image-base approach \cite{sinha2016deep}, by a large margin, and comes close to models such as VoxNet \cite{maturana2015voxnet}.

\begin{figure}[ht!]
  \centering
  \begin{subfigure}[b]{0.2\textwidth}
    \centering
    \includegraphics[width=\linewidth]{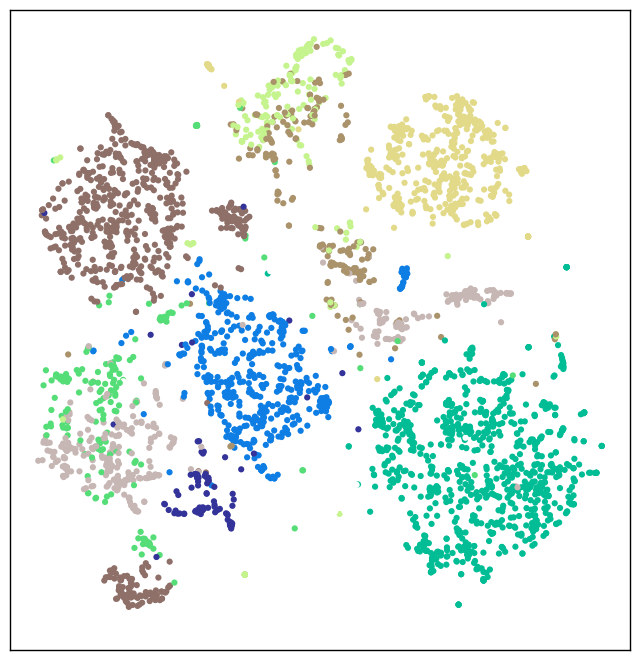}
    \caption*{\it ModelNet10}
  \end{subfigure}
  \qquad
  \begin{subfigure}[b]{0.2\textwidth}
    \centering
    \includegraphics[width=\linewidth]{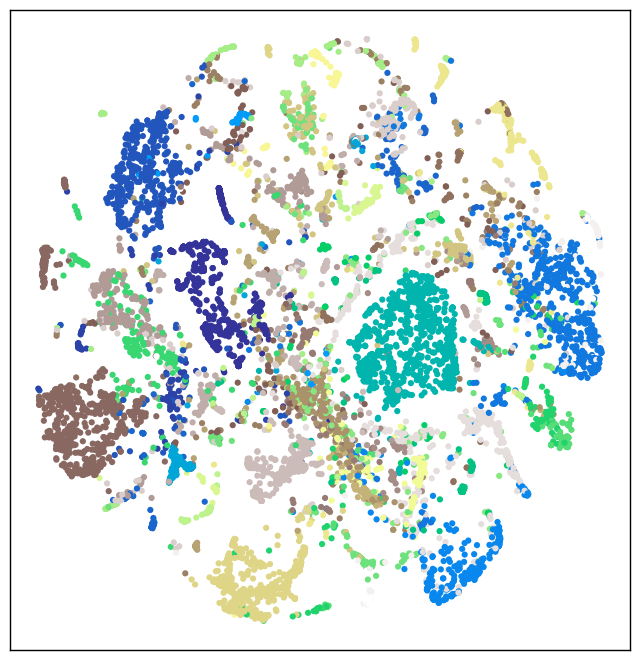}
    \caption*{\it ModelNet40}
  \end{subfigure}
  \caption{t-SNE plots of the latent embeddings for ModelNet10 and ModelNet40. Each color represents one class.}
  \label{fig:tsne_viz}
\end{figure}

In order to visualise the learned feature embeddings, we employ t-SNE \cite{maaten2008visualizing} to map our high dimensional feature to a 2D plane. The visualization is shown in Figure \ref{fig:tsne_viz}.

\vspace{-0.5em}

\subsection{Single Image 3D Model Retrieval}
\label{subsec:imrec}
Real-world, single image 3D model retrieval is another application of the proposed VSL model. This is a challenging problem, forcing a model to deal with real-world 2D images under a variety of lighting conditions and resolutions. Furthermore, there are many instances of model occlusion as well as different colour gradings.

\begin{table*}[ht!]
  \centering
  \small
  \setlength\tabcolsep{6pt}
    \begin{tabular}{|c||c|c|c|c|c|c|c|c|c|c||c|}
      \hline
     & {\bf aero}  & {\bf bike}  & {\bf boat}  & {\bf bus}   & {\bf car}   & {\bf chair} & {\bf mbike} & {\bf sofa } & {\bf train} & {\bf tv}    & {\bf mean} \\
      \hline
    NRSfM & 0.298 & 0.144 & 0.188 & 0.501 & 0.472 & 0.234 & 0.361 & 0.149 & 0.249 & 0.492 & 0.318 \\
      \hline
    3D-R2N2 [LSTM-1] & 0.472 & 0.330  & 0.466 & 0.677 & 0.579 & 0.203 & 0.474 & 0.251 & 0.518 & 0.438 & 0.456 \\
      \hline
      3D-R2N2 [Res3D-GRU-3] & 0.544 & 0.499 & {\bf 0.560}  & 0.816 & 0.699 & 0.280  & 0.649 & 0.332 & 0.672 & {\bf 0.574} & 0.571 \\
      \hline
      VSL (jointly trained) & 0.514 & 0.269 & 0.327  & 0.558 & 0.633 & 0.199  & 0.301 & 0.173 & 0.402 & 0.337 & 0.432 \\
      \hline
      VSL (separately trained) & {\bf 0.631} & {\bf 0.657} & 0.554  & {\bf 0.856} & {\bf 0.786} & {\bf 0.311}  & {\bf 0.656} & {\bf 0.601} & {\bf 0.804} & 0.454 & {\bf 0.619} \\
    \hline
    \end{tabular}%
  \caption{Per-category voxel predictive performance on PASCAL 3D, as measured by Intersection-of-Union (IoU).}
  \label{tab:ioucompare}
\end{table*}

To test our model on this application, we use the PASCAL 3D \cite{xiang_wacv14} dataset and utilize the same exact training and testing splits from \cite{kar2015category}. We compare our results with those reported for recent approaches, including the NRSfM \cite{kar2015category} and 3D-R2N2 \cite{choy20163d} models. Note that these also used the exact same experimental configurations we did.

\begin{figure}[ht!]
  \centering
  \small
  \setlength\tabcolsep{3pt}
    \begin{tabular}{ccccc}
    Input & GT & VSL & 3D-R2N2\cite{choy20163d}   & NRSfM\cite{kar2015category}  \\
    \toprule
   \includegraphics[width=0.16\linewidth]{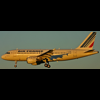}    &  \includegraphics[width=0.16\linewidth]{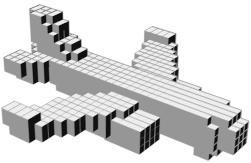}     &  \includegraphics[width=0.16\linewidth]{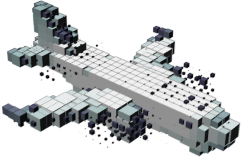}     &    \includegraphics[width=0.16\linewidth]{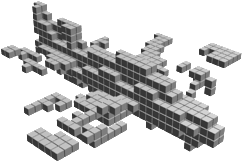}   &    \includegraphics[width=0.16\linewidth]{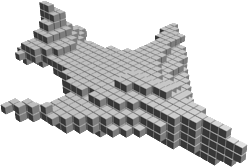}   \\
    \includegraphics[width=0.16\linewidth]{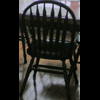}    &  \includegraphics[height=0.16\linewidth]{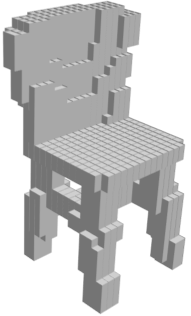}     &  \includegraphics[height=0.16\linewidth]{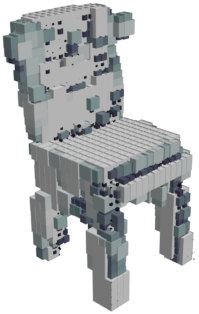}    &    \includegraphics[height=0.16\linewidth]{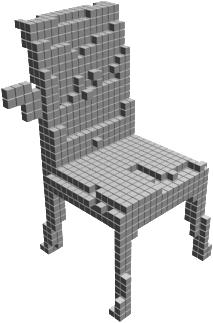} &    \includegraphics[height=0.16\linewidth]{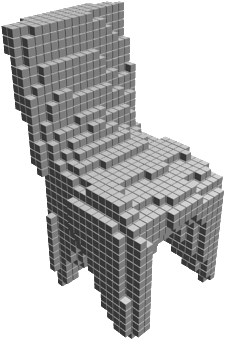}      \\
    \includegraphics[width=0.16\linewidth]{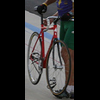}     &  \includegraphics[width=0.16\linewidth]{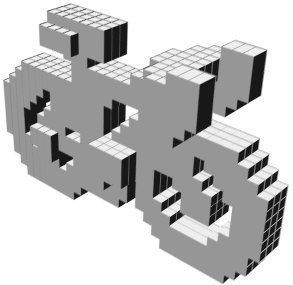}   &    \includegraphics[width=0.16\linewidth]{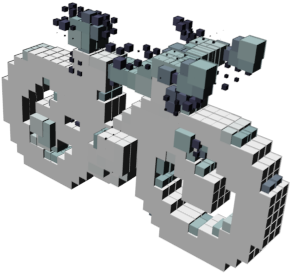}      &   \includegraphics[width=0.16\linewidth]{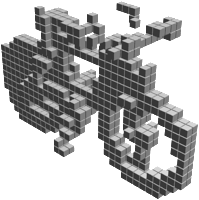}     & \includegraphics[width=0.16\linewidth]{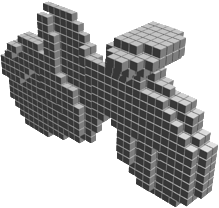}      \\  \includegraphics[width=0.16\linewidth]{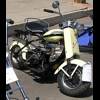}     &  \includegraphics[width=0.16\linewidth]{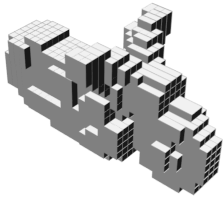}   &    \includegraphics[width=0.16\linewidth]{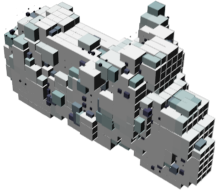}      &   \includegraphics[width=0.16\linewidth]{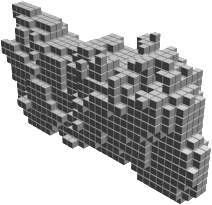}       & \includegraphics[width=0.16\linewidth]{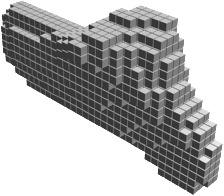}   \\
    \includegraphics[width=0.16\linewidth]{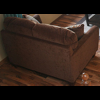}    &  \includegraphics[width=0.16\linewidth]{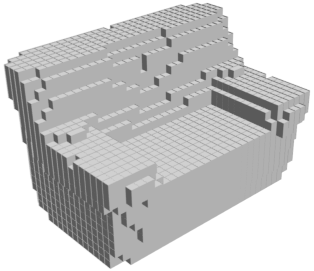}     &  \includegraphics[width=0.16\linewidth]{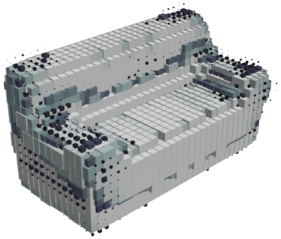}     &    \includegraphics[width=0.16\linewidth]{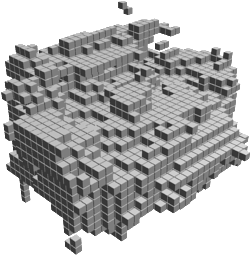}   &    \includegraphics[width=0.16\linewidth]{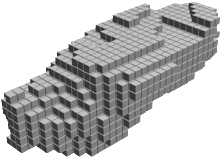}   \\ \includegraphics[width=0.16\linewidth]{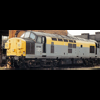}     &  \includegraphics[width=0.16\linewidth]{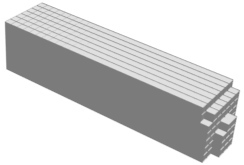}   &    \includegraphics[width=0.16\linewidth]{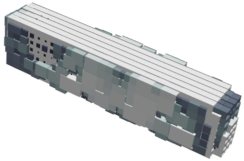}      &   \includegraphics[width=0.16\linewidth]{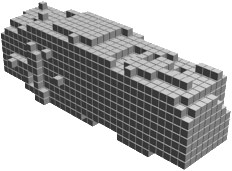}       & \includegraphics[width=0.16\linewidth]{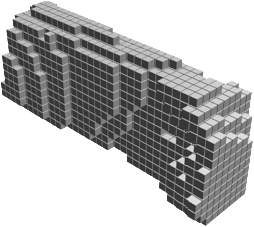}  \\
    \end{tabular}%
  \caption{Reconstruction samples for PASCAL 3D from the separately trained VSL.}
  \label{fig:imrec}
  \vspace{-1em}
\end{figure}

For this task, we train our model in two different ways: 1) jointly on all categories, and 2) separately on each category. In Figure \ref{fig:imrec}, we observe better reconstructions from the (separately-trained) VSL when compared to previous work. Unlike the NRSfM \cite{kar2015category}, the VSL does not require any segmentation, pose information, or keypoints. In addition, the VSL is trained from scratch while the 3D-R2N2 is pre-trained using the ShapeNet dataset \cite{chang2015shapenet}. However, the jointly-trained VSL did not outperform the 3D-R2N2, which is also jointly-trained. The performance gap is due to the fact that the 3D-R2N2 is specifically designed for image reconstruction and employs a residual network \cite{he2016deep} to help the model learn richer semantic features.

Quantitatively, we compare our VSL to the NRSfM \cite{kar2015category} and two versions of 3D-R2N2 from \cite{choy20163d}, one with an LSTM structure and another with a deep residual network. Results (Intersection-of-Union) are shown in Table \ref{tab:ioucompare}. Observe that our jointly trained model performs comparably to the 3D-R2N2 LSTM variant while the separately trained version surpasses the 3D-R2N2 ResNet structure in 8 out of 10 categories, half of them by a wide margin.

\subsection{Shape Arithmetic}
\label{subsec:shapeplus}
\begin{figure}[ht!]
  \vspace{-0.75em}
  \centering
  \includegraphics[width=\linewidth]{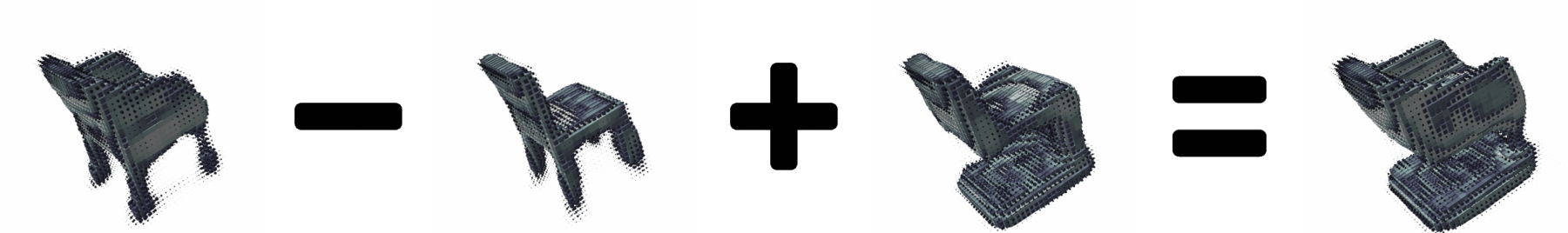}  \\
  \includegraphics[width=\linewidth]{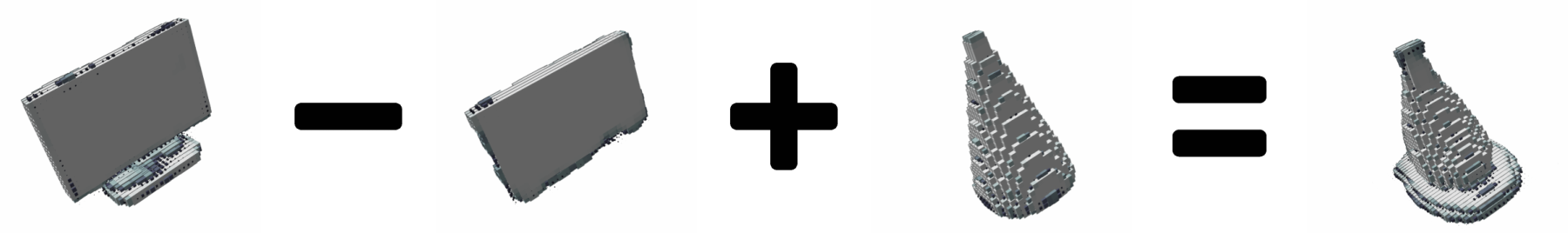}  \\
  \includegraphics[width=\linewidth]{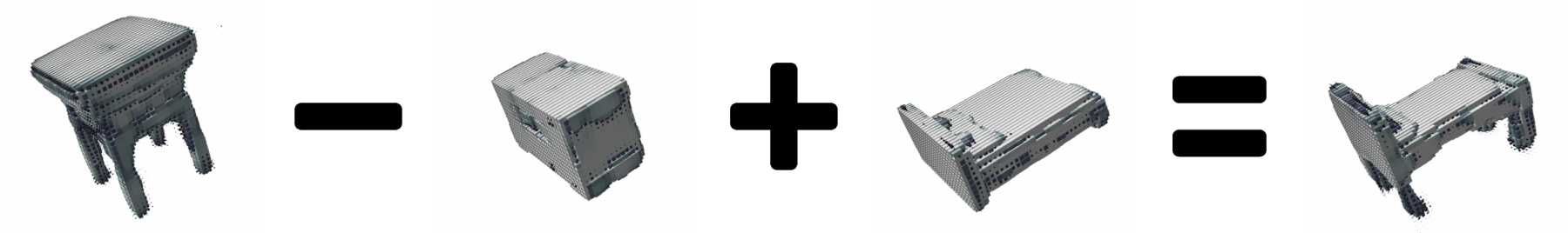}  \\
  \includegraphics[width=\linewidth]{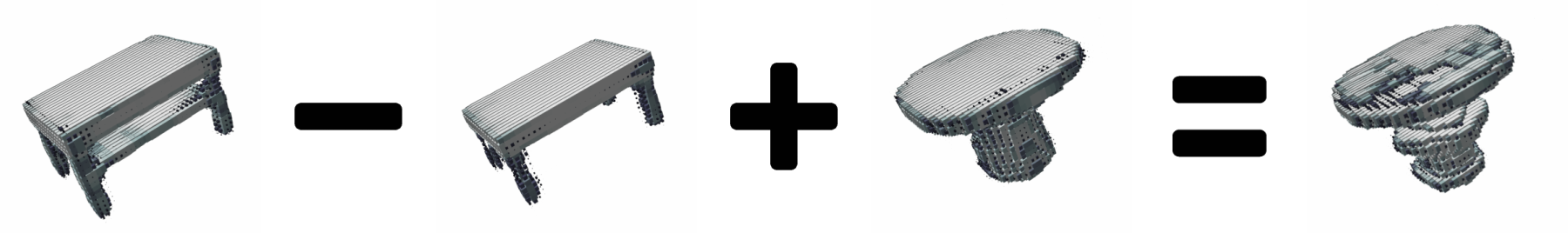}  \\
  \caption{Samples of our shape arithmetic experiment.}
  \label{fig:shapearith}
  \vspace{-0.75em}
\end{figure}

Another way to explore the learned embeddings is to perform various vector operations on the latent space, much what was done in \cite{wu2016learning,girdhar2016learning}. We present some results of our shape arithmetic experiment in Figure \ref{fig:shapearith}. Different from previous results, all of our objects are sampled from the model embeddings which were trained using the whole dataset with 40 classes. Furthermore, unlike the blurrier generations of \cite{girdhar2016learning}, the VSL seems to generate very interesting combinations of the input embeddings without the need for any matching to actual 3D shapes from the original dataset. The resultant objects appear to clearly embody the intuitive meaning of the vector operators.

\section{Conclusion}
\label{chp:conclusion}
In this paper, we proposed the Variational Shape Learner, a hierarchical latent-variable model for 3D shape modelling, learnable through variational inference. In particular, we have demonstrated 3D shape generation results on a popular benchmark, the ModelNet dataset. We also used the learned embeddings of our model to obtain state-of-the-art in unsupervised shape classification and furthermore showed that we could generate unseen shapes using shape arithmetic. Future work will entail a more thorough investigation of the embeddings learned by our hierarchical latent-variable model as well as integration of better prior distributions into the framework.

{\small
\bibliographystyle{ieee}
\bibliography{egbib}
}

\end{document}